\pdfoutput=1

\documentclass[11pt]{article}

\usepackage[final]{acl}  

\usepackage{times}
\usepackage{latexsym}

\usepackage[T1]{fontenc}

\usepackage[utf8]{inputenc}

\usepackage{microtype}
\usepackage{booktabs}

\usepackage{inconsolata}

\usepackage{graphicx}

\title{Ustnlp16 at SemEval-2025 Task 9: Improving Model Performance through Imbalance Handling and Focal Loss}

\author{
  \textbf{Zhuoang Cai}, \textbf{Zhenghao Li}, \textbf{Yang Liu}, \textbf{Liyuan Guo}, \textbf{Yangqiu Song} \\
  Department of Computer Science and Engineering, \\
  The Hong Kong University of Science and Technology (HKUST), Hong Kong SAR \\
  \texttt{\{zcaiat, zlika, yliuhy, lguoar\}@connect.ust.hk, yqsong@cse.ust.hk}
}
\usepackage{amsmath}
\usepackage{tikz}
\usetikzlibrary{arrows,shapes,positioning}

\begin{document}
\maketitle
\begin{abstract}
Classification tasks often suffer from imbalanced data distribution, which presents challenges in food hazard detection due to severe class imbalances, short and unstructured text, and overlapping semantic categories. In this paper, we present our system for SemEval-2025 Task 9: Food Hazard Detection, which addresses these issues by applying data augmentation techniques to improve classification performance. We utilize transformer-based models, BERT and RoBERTa, as backbone classifiers and explore various data balancing strategies, including random oversampling, Easy Data Augmentation (EDA), and focal loss. Our experiments show that EDA effectively mitigates class imbalance, leading to significant improvements in accuracy and F1 scores. Furthermore, combining focal loss with oversampling and EDA further enhances model robustness, particularly for hard-to-classify examples. These findings contribute to the development of more effective NLP-based classification models for food hazard detection.
\end{abstract}

\section{Introduction}

The rapid advancement of natural language processing (NLP) has facilitated the development of automated classification systems across various domains, including food hazard detection. Accurately identifying and categorizing food hazards is essential for ensuring food safety and mitigating health risks associated with contaminated or unsafe food products. However, food hazard classification presents several challenges, including severe class imbalances, ambiguous and unstructured textual descriptions, and the need for high predictive accuracy. Traditional approaches to hazard detection have relied on manual inspections and rule-based classification methods, which are often time-consuming and prone to human error. In contrast, recent advancements in machine learning and NLP have enabled the automation of this process, leveraging text classification models to analyze food hazard reports and categorize them into predefined classes.

Transformer-based models, such as BERT \cite{devlin2018bert} and RoBERTa \cite{zhuang-etal-2021-robustly}, have demonstrated state-of-the-art performance in text classification tasks. However, their effectiveness in imbalanced classification settings remains a challenge, as they tend to favor majority classes while underperforming in minority categories. Class imbalance is a common issue where certain categories have significantly fewer instances than others, leading to biased predictions and reduced model performance on underrepresented classes. To address this issue, researchers have explored various techniques, including data augmentation, resampling methods, and modified loss functions. Easy Data Augmentation (EDA) \cite{wei2019eda} generates additional training samples for minority classes, enhancing model generalization. Similarly, focal loss \cite{lin2017focal} modifies the traditional cross-entropy loss function to focus more on difficult-to-classify examples, improving performance on underrepresented categories.

In this study, we systematically investigate the impact of data balancing techniques on transformer-based models for food hazard classification. Specifically, we evaluate the effectiveness of oversampling, EDA, and focal loss in mitigating class imbalance and improving classification performance. Through extensive experimentation, we demonstrate that these strategies enhance model robustness, particularly in detecting minority-class hazards. Our findings contribute to the development of more reliable NLP-based classification models for food safety applications, providing valuable insights into optimal approaches for handling class imbalance in text classification tasks.

\section{Related Work}
\subsection{Text Classification}
Text classification, a core NLP task, involves assigning predefined labels to text. Traditional methods used rule-based approaches and machine learning models like Naïve Bayes, SVM, and Random Forests. Deep learning, particularly transformer-based models, has significantly improved performance by capturing contextual dependencies~\cite{fan2024chainofchoicehierarchicalpolicylearning}.

Models like BERT \cite{devlin2018bert} and RoBERTa \cite{zhuang-etal-2021-robustly} set new benchmarks but struggle with imbalanced datasets, where minority classes are often overlooked. This issue is critical in domains like food hazard detection, where rare cases carry significant risks. To address class imbalance, researchers employ resampling techniques \cite{lauron2016improved}, cost-sensitive learning, and modified loss functions like focal loss \cite{lin2017focal}. Data augmentation has also proven effective in enhancing classification robustness, especially in low-resource settings.

\subsection{Data Augmentation}
Data augmentation expands training datasets to improve model generalization~\cite{fan-etal-2024-goldcoin}, particularly in NLP, where it helps mitigate class imbalance in text classification. Traditional methods like synonym replacement, back-translation, and paraphrasing \cite{wei2019eda} enhance lexical diversity while preserving meaning. Easy Data Augmentation (EDA) is widely used due to its simplicity, applying synonym replacement, random insertion, swap, and deletion to boost performance on imbalanced datasets. More advanced techniques leverage contextual embeddings (e.g., Word2Vec, FastText) and transformer-based text generation, though excessive alterations risk label noise. Recent studies combine augmentation with resampling strategies \cite{foods13203300}, improving accuracy and F1 scores, especially for underrepresented classes. Building on these advancements, we integrate oversampling, EDA, and focal loss to enhance classification in food hazard detection. Our approach effectively mitigates class imbalance and strengthens model robustness, particularly for minority classes.

\section{Task Definition and Dataset}

SemEval-2025 Task 9 involves classifying short food recall reports into predefined hazard-related labels \cite{semeval2025-task9}. The objective is to develop robust NLP models that accurately identify hazard types and product categories despite imbalanced, noisy, and unstructured text.

The dataset comprises thousands of entries obtained from government agencies. Each entry includes a title, typically a brief recall identifier, and a text description that varies in length and format, often containing domain-specific terminology. The data are marked by severe class imbalance, with many hazard categories significantly underrepresented.

\section{Methodology}

Recent work in food hazard detection highlights the importance of addressing data imbalance, short and unstructured text, and overlapping semantic categories \cite{foods13203300}. SemEval-2025 Task 9 intensifies these challenges by providing a real-world dataset where certain hazard classes are severely underrepresented, necessitating specialized techniques to ensure fair and robust classification. In this study, we adopt a transformer-based approach, leveraging BERT \cite{devlin2018bert} and RoBERTa \cite{zhuang-etal-2021-robustly}, and integrate three key strategies—random oversampling, Easy Data Augmentation (EDA), and focal loss—to enhance performance on minority classes while maintaining overall accuracy.

\begin{figure}
    \centering
    \includegraphics[width=\linewidth]{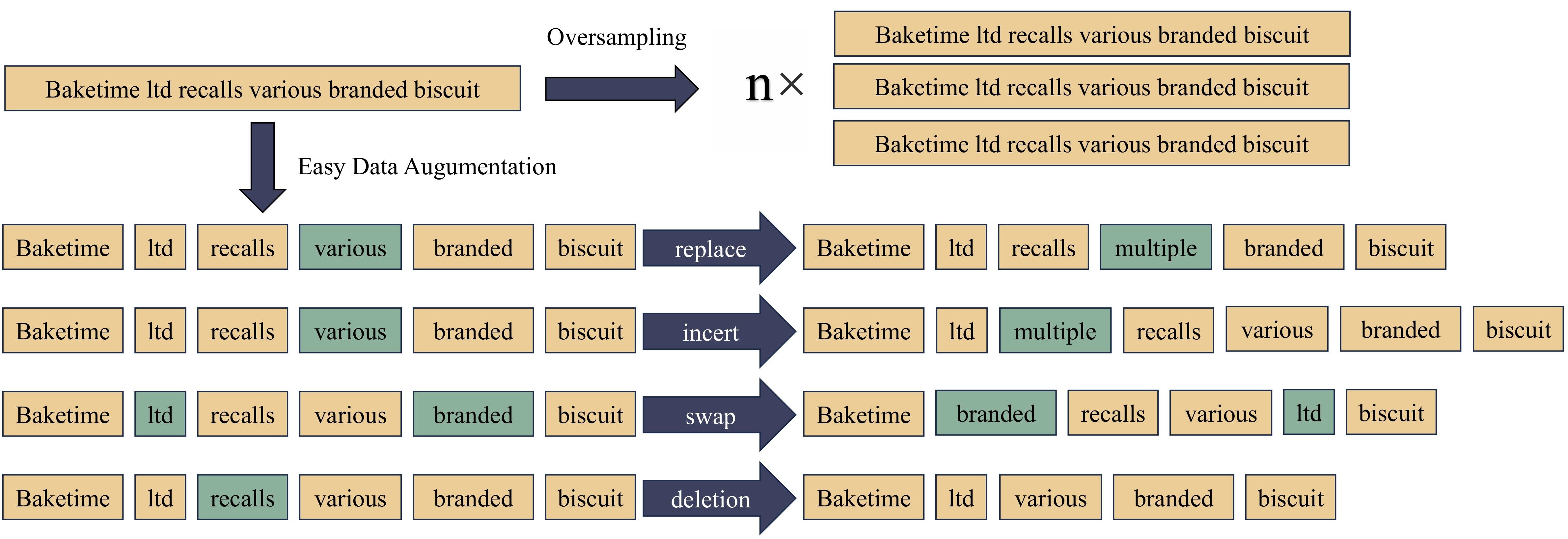}
    \caption{The structured pipeline for food hazard detection.}
    \label{fig:enter-label}
\end{figure}

Our methodology follows a structured pipeline, shown in Figure~\ref{fig:enter-label}, which consists of data preprocessing and augmentation, transformer-based model fine-tuning, and final evaluation using standard classification metrics. The sections below describe how these components are cohesively combined to tackle the real-world complexities of food hazard reports.

\subsection{Data Preprocessing and Augmentation}

All textual inputs undergo cleaning and tokenization before fine-tuning. We remove stopwords, numerical tokens, and other non-informative elements, followed by lemmatization to standardize word forms. Outliers are then filtered using the interquartile range (IQR) to reduce extreme text lengths that could bias the model. We adopt WordPiece tokenization \cite{song2020fast} to handle out-of-vocabulary (OOV) tokens, thereby preserving subword-level information crucial for short and domain-specific texts.

To counteract severe class imbalance, we introduce a combined augmentation strategy that integrates random oversampling with Easy Data Augmentation (EDA) \cite{wei2019eda}. Oversampling is performed after tokenization to ensure each minority class is represented at a target sample rate. EDA, summarized in Table~\ref{tab:eda_operations}, is then applied to further expand the diversity of minority samples by introducing lexical and structural variations. Rather than applying each augmentation technique independently, we incorporate them into a unified process that consistently enhances minority-class coverage and lexical variety. This integrated augmentation stage aligns with prior findings that emphasize synergy between resampling and data augmentation for imbalanced text classification \cite{lauron2016improved}.

\begin{table}[t]
\centering
\renewcommand{\arraystretch}{1.2}
\setlength{\tabcolsep}{4pt}
\resizebox{\linewidth}{!}{%
\begin{tabular}{p{3.8cm} p{4cm}}
\toprule
\textbf{Operation} & \textbf{Description} \\
\midrule
Synonym Replacement & Replace with synonyms \\
Random Insertion & Insert random words \\
Random Swap & Swap two words \\
Random Deletion & Remove words (\( p = 0.1 \)) \\
\bottomrule
\end{tabular}%
}
\caption{Easy Data Augmentation (EDA) operations applied to the dataset.}
\label{tab:eda_operations}
\end{table}

\subsection{Classification Model and Imbalance Handling}

The classification model builds on BERT and RoBERTa, which are fine-tuned for multi-class prediction. While cross-entropy loss remains the baseline choice, we adopt focal loss \cite{lin2017focal} to emphasize hard-to-classify examples in minority classes. The focal loss function is given by:
\begin{equation}
FL(p_t) = -\alpha_t (1 - p_t)^\gamma \log(p_t),
\end{equation}
where \( \alpha_t \) balances class contributions, \( \gamma \) focuses on difficult samples, and \( p_t \) is the predicted probability for the correct class. We set \(\alpha=1\) and \(\gamma=2\) based on initial experiments indicating improved recall for underrepresented hazards.

Random oversampling is performed using the strategy:
\begin{equation}
\begin{aligned}
\text{sampling\_strategy} &= \{\, k : \text{target\_count} \;\mid \\
&\quad v < \text{target\_count}, \forall k, v \},
\end{aligned}
\end{equation}
where each minority class is upsampled to match a threshold of the majority class size. By applying oversampling in tandem with EDA, we ensure that minority classes benefit from both quantitative and qualitative increases in training samples.

\subsection{System Configurations}

We evaluate several configurations to highlight the effect of each balancing technique (Table~\ref{tab:system_variants}). The \textit{Baseline} employs standard BERT fine-tuning without augmentation, while additional setups incorporate oversampling, EDA, focal loss, or a combination thereof. We also include RoBERTa variants, reflecting the same augmentation and imbalance strategies. This design enables a comprehensive comparison of how each technique—alone or combined—contributes to classification performance on short, imbalanced food hazard reports.

\begin{table*}[t]
\small
\centering
\renewcommand{\arraystretch}{1.2}
\setlength{\tabcolsep}{5pt}
\begin{tabular}{p{6.2cm}p{8cm}}
\toprule
\textbf{Configuration} & \textbf{Description} \\
\midrule
Baseline & Standard BERT fine-tuning with cross-entropy loss \\
BERT + Oversampling & Resample minority classes after tokenization \\
BERT + EDA & Apply data augmentation using EDA \\
BERT + Focal Loss & Replace cross-entropy with focal loss \\
BERT + EDA + Focal Loss & Combine lexical augmentation and focal loss \\
RoBERTa Variants & Mirror each configuration using RoBERTa \\
\bottomrule
\end{tabular}
\caption{Model configurations used for evaluation.}
\label{tab:system_variants}
\end{table*}

This integrated methodology ensures that each stage—preprocessing, augmentation, model training—cooperates to address the unique challenges posed by SemEval-2025 Task 9, namely short, imbalanced, and domain-specific textual data.

\section{Experiments}
Our study strategically integrates three techniques—Easy Data Augmentation (EDA), oversampling and focal loss—to address class imbalance in classification tasks. The sequence of application is as follows: EDA is applied before tokenization to enhance data diversity, oversampling is applied after tokenization to balance class distribution, and focal loss is utilized during training to optimize the model's focus on difficult samples. These experiments not only demonstrate the effectiveness of each individual method but also highlight the synergistic benefits of their combination. The results show that this integrated approach enhances data diversity, balances class distribution, and improves model performance by prioritizing challenging samples.
\subsection{Experimental Setup}
\paragraph{Oversampling}
The sample rate is set to r. For classes whose size is smaller than r\% of the most-frequent class, we perform oversampling to ensure their size matches that of r\% of the most-frequent class.

\paragraph{Easy Data Augmentation}
The sample rate is set to r. For each instance input, we perform EDA. Each of the four operations—Random Synonym Replacement, Random Insertion, Random Swap, and Random Deletion—has a 50\% probability of being applied. For the first three operations, the parameter (n), which indicates the number of times the operation is to be performed, is randomly selected within the range of 1 to the total number of words in the text. In the case of the Random Deletion operation, each word is assigned a 10\% probability of being deleted.

\paragraph{Focal Loss}
Alpha ($\alpha$),the balance parameter for class imbalance, is set to 1. Gamma ($\gamma$), the focusing parameter for hard examples, is set to 2. 
The method for aggregating the loss values (reduction) is "mean".
\subsection{Training Details}
We utilized the 'title' and 'text' fields from the dataset released by the organizers. In the data processing phase, categorical labels were encoded into numerical values using the LabelEncoder. The dataset was subsequently split into training and testing sets, with 20\% allocated for testing. If Easy Data Augmen-
tation (EDA) was enabled, data augmentation techniques were applied specifically to the training subset (train\_df). Additionally, if oversampling was employed, data augmentation was conducted after the tokenization process.

For our models, we utilized BERT \cite{devlin2018bert} and RoBERTa \cite{zhuang-etal-2021-robustly}. During training, the total batch size was set to 32. The AdamW optimizer \cite{kingma2014adam} was used with a learning rate of 5e-5, and dropout was specified at 0.0. The learning rate schedule followed a 'cosine with warmup' strategy, incorporating a warmup phase equivalent to 10\% of the total training steps to gradually adjust the learning rate and enhance model convergence. The default loss function used was CrossEntropyLoss, unless focal loss was selected.

\begin{table*}[t]
    \small
    \centering
    \begin{tabular}{l c c c}
        \multicolumn{4}{c}{\textbf{(ST2) Product Detection}} \\
        \hline
        \textbf{Training Methods} & \textbf{Accuracy} & \textbf{F1-Macro} & \textbf{F1-Weighted} \\
        \hline
        $\mathrm{BERT}_{base}$ & 0.22 & 0.03 & 0.13 \\
        $\mathrm{Oversampling}_{0.1}$ & 0.50 & 0.25 & 0.45 \\
        $\mathrm{Oversampling}_{0.2}$ & 0.45 & 0.22 & 0.41 \\
        $\mathrm{Oversampling}_{0.5}$ & 0.47 & 0.23 & 0.42 \\
        $\mathrm{Oversampling}_{1.0}$ & 0.29 & 0.13 & 0.26 \\
        $\mathrm{EDA}_{0.1}$ & 0.54 & 0.29 & 0.50 \\
        $\mathrm{EDA}_{0.2}$ & 0.55 & 0.30 & 0.52 \\
        $\mathrm{EDA}_{0.5}$ & 0.55 & 0.30 & 0.51 \\
        $\mathrm{EDA}_{1.0}$ & 0.54 & 0.30 & 0.51 \\
        $\mathrm{Focal\ loss + Oversampling}_{0.1}$ & 0.49 & 0.24 & 0.44 \\
        $\mathrm{Focal\ loss + Oversampling}_{0.2}$ & 0.47 & 0.23 & 0.41 \\
        $\mathrm{Focal\ loss + Oversampling}_{0.5}$ & 0.48 & 0.25 & 0.44 \\
        $\mathrm{Focal\ loss + EDA}_{0.1}$ & 0.53 & 0.29 & 0.50 \\
        $\mathrm{Focal\ loss + EDA}_{0.2}$ & 0.54 & 0.29 & 0.51 \\
        $\mathrm{Focal\ loss + EDA}_{0.5}$ & 0.54 & 0.30 & 0.51 \\
        $\mathrm{Focal\ loss + EDA}_{1.0}$ & 0.53 & 0.29 & 0.50 \\
        $\mathrm{Oversampling \ + EDA}_{0.1}$ & 0.49 & 0.25 & 0.45 \\
        \hline
    \end{tabular}
    \caption{\label{product}
        BERT model with different strategies for predicting product. The subscript isthe  sample rate. For example, 0.1 means to upsample categories that are less than 10\% of the maximum sample to 10\% of the maximum sample.
    }
\end{table*}

\begin{table*}[t]
  \centering
  \small
  \begin{tabular}{lccc}
    \multicolumn{4}{c}{\textbf{(ST2) Hazard Detection}} \\
    \hline
    \textbf{Training Methods}           & \textbf{Accuracy} & \textbf{F1-Macro} & \textbf{F1-Weighted}\\
    \hline
    $\mathrm{BERT}_{base}$ & 0.58 & 0.17 & 0.53 \\
    $\mathrm{BERT \ +Focal\ loss + EDA}_{0.1}$ & 0.86 & 0.59 & 0.85 \\
    $\mathrm{RoBERTa \ +Focal\ loss + EDA}_{0.1}$ & 0.86 & 0.59 & 0.85 \\
    \hline
  \end{tabular}
  \caption{\label{hazard}
    BERT and RoBERTa with focal loss and EDA on predicting hazard. The subscript is sample rate. For example, 0.1 means to upsample categories that are less than 10\% of the maximum sample to 10\% of the maximum sample.
  }
\end{table*}
In our study on NLP food hazard classification, we concentrated on addressing the issue of class imbalance, particularly in predicting the most imbalanced label, "product." To evaluate our model's performance, we established BERT, provided by the organizers, as our baseline. The evaluation utilized inputs from both "text" and "title" to enhance the model's effectiveness. We employed several metrics to assess performance, including accuracy, F1-score (macro), and F1-score (weighted). See Table~\ref{product} for the results for predicting "product" (data split: training/validation/test).
\paragraph{Oversampling}
BERT achieved an accuracy of \(0.22\), with an F1-Macro score of \(0.03\) and an F1-Weighted score of \(0.13\), highlighting its limitations in handling the imbalanced dataset. Oversampling with a sample rate at \(0.1\) yielded the best performance among the oversampling variations, achieving an accuracy of \(0.50\), an F1-Macro score of \(0.25\), and an F1-Weighted score of \(0.45\). This indicates that oversampling can effectively address class imbalance and improve classification performance.
\paragraph{Easy Data Augmentation}
The experimental results presented in Table~\ref{product} also demonstrate the effectiveness of Easy Data Augmentation (EDA) in enhancing the performance of the model for the ``product.'' Compared to the baseline accuracy of \(0.22\), the application of EDA across multiple configurations consistently improved performance. EDA with a sample rate of \(0.2\) increased accuracy to a maximum of \(0.55\). Similarly, the F1 macro score improved significantly from \(0.03\) to \(0.30\), while the F1 weighted score rose from \(0.13\) to \(0.52\). These findings underscore the potential of EDA as a valuable technique for improving model performance in imbalanced classification tasks.
\paragraph{Combination}
After observing that handling imbalance can enhance model performance, we explored the effects of combining this strategy with focal loss, which places greater emphasis on harder-to-classify examples. We also examined the combination of EDA with oversampling. While these combinations did result in performance improvements that surpassed the baseline BERT model, they did not achieve the same high levels of effectiveness as EDA alone. For instance, the combination of focal loss with oversampling yielded an accuracy of \(0.49\) at a sample rate of \(0.1\), while EDA at the same rate achieved an accuracy of \(0.54\).

\paragraph{Prediction on Hazard}
To ensure that the model predicts well across different tasks, we also made predictions on ``hazard.'' The combination of focal loss and Easy Data Augmentation (EDA) significantly increased performance from an accuracy of \(0.58\) for BERT to \(0.86\) at a sample rate of \(0.1\). We also investigated the RoBERTa model in a similar manner; however, it performed comparably to BERT, leading us to discontinue further exploration of RoBERTa. Table~\ref{hazard} shows some results of these experiments.

In summary, the results show that this integrated approach enhances data diversity, balances class distribution, and improves model performance by prioritizing challenging samples. Our findings indicate that EDA is a particularly effective technique for addressing class imbalance in food hazard classification. While combinations with focal loss and oversampling improve performance, they do not surpass EDA alone. Additionally, these methods may require further fine-tuning and more training steps to optimize their effectiveness.

\section{Conclusion}
In this paper, we strategically combined three methods - Easy Data Augmentation (EDA), oversampling, and focal loss - to tackle class imbalance in classification tasks. Our model ranked 13th on ST1, and 12th on ST2. It surpassed the baseline in both ST1, predicting product and hazard category, and ST2, predicting specific hazard and product.

Future work will explore strategies to enhance model performance in data-constrained and domain-shift scenarios. We plan to investigate the integration of advanced augmentation techniques and refined loss functions, along with fine-tuning the existing methodology.

\newpage
\section*{Limitations}
While our approach demonstrated notable improvements in food hazard classification, several limitations remain. First, despite the effectiveness of EDA and oversampling in mitigating class imbalance, these techniques may introduce synthetic noise into the dataset. Augmented samples, particularly those generated via lexical modifications, may not always accurately preserve the semantic meaning of the original text, potentially leading to misclassification.

Second, our reliance on transformer-based models presents computational challenges. Fine-tuning these models requires significant hardware resources, making it less feasible for real-time applications or deployment in resource-constrained environments. Additionally, while focal loss improves performance on hard-to-classify examples, it requires careful tuning of hyperparameters, which may not generalize well across different datasets or classification tasks.

Another key limitation is the potential lack of generalizability. Our model was trained on the SemEval-2025 Task 9 dataset, which, despite being a real-world dataset, has specific linguistic characteristics and class distributions. This may limit the model’s ability to perform well on other food safety-related classification tasks with different text structures, hazard categories, or reporting styles. Future work should explore domain adaptation techniques and evaluate performance across multiple datasets.

Finally, while our approach improves minority class detection, the gap between majority and minority class performance remains. Additional techniques, such as contrastive learning, cost-sensitive training, or adaptive resampling, could be explored to further enhance model fairness and robustness.

\newpage
\bibliography{custom}

\end{document}